# یافتن ویژگی‌های موثر با استفاده از رویکرد ابرابتکاری

میترا منتظری[1]*؛ مهدیه سلیمانی باغشاه[2]؛ علی اکبر نیک‌نفس[3]

**چکیده**

با پیدایش پایگاه داده‌های بزرگ و درکنار آن نیاز به روش‌های یادگیری مناسب مشکلات جدیدی پا به عرصه گذاشته‌اند که باعث به وجود آمدن راه‌کارهایی برای انتخاب ویژگی‌های موثر پایگاه داده‌ها شده‌اند. انتخاب ویژگی، مساله یافتن ویژگی‌های موثر از میان ویژگی‌های موجود است، به طوری که مجموعه‌ی حاصل باعث افزایش دقت و کاهش پیچیدگی گردد. برای تشخیص این که کدام زیرمجموعه موثرتر است، یک راه‌حل بررسی تمامی زیرمجموعه‌های ممکن است. اما از آن‌جاکه بررسی همه حالات جزء مسایل سخت و دارای پیچیدگی محاساباتی بالا است، تاکنون الگوریتم‌های جستجوی ابتکاری متعددی برای این منظور معرفی شده است. رویکرد ابرابتکاری یک رویکرد جدید جستجو است که می‌تواند فضای جستجو را با به کاربستن چند جستجوی محلی، که هر یک پیمایشگرهایی در همسایگی راه حل هستند، به طور کارآمد جستجو کند. با توجه به اینکه هر ناحیه از فضای جستجو ویژگی‌های خاص خود را دارد، در مسیر جستجو بایستی یک جستجوی محلی مناسب انتخاب و در راه‌حل جاری به کاربسته شود. این انتخاب به عهده‌ی یک ناظر است. ناظر در هر زمان، انتخاب را بر اساس تاریخچه عملکرد جستجوهای محلی انجام می‌دهد. این رویکرد با انجام چنین تکنیکی می‌تواند هم کاوش و هم بهره‌وری را به خوبی انجام دهد. الگوریتم‌های ابتکاری موجود نمی توانستند مصالحه خوبی بین کاوش و بهره‌وری داشته باشند در نتیجه فضای جستجو به خوبی جستجو نمی‌گردید و سرعت همگرایی پایینی داشتند.

در این مقاله برای اولین بار از رویکرد ابرابتکاری جهت پیداکردن ویژگی‌های موثر برای دسته‌بندی استفاده شده است. در الگوریتم پیشنهادی، الگوریتم ژنتیک به عنوان یک ناظر به‌کار گرفته شده و ۱۶ الگوریتم ابتکاری به منظور جستجوی محلی تعریف شده‌است. نتایج حاصل از اعمال الگوریتم پیشنهادی بر روی پایگاه داده‌های گرفته شده از UCI نشان دهنده کارآمد بودن الگوریتم پیشنهادی در مقایسه با روش‌های ابتکاری مطرح موجود برای انتخاب ویژگی است.

**کلمات کلیدی**

انتخاب ویژگی، رویکرد ابرابتکاری، الگوریتم‌های فراابتکاری، الگوریتم‌های ممتیک، جستجوی محلی.

# Selecting Efficient Features via a Hyper-Heuristic Approach

Mitra Montazei; Mahdieh Soleymani Baghshah; Aliakbar Niknafs

**ABSTRACT**

By Emerging huge databases and the need to efficient learning algorithms on these datasets, new problems have appeared and some methods have been proposed to solve these problems by selecting efficient features. Feature selection is a problem of finding efficient features among all features in which the final feature set can improve accuracy and reduce complexity. One way to solve this problem is to evaluate all possible feature subsets. However, evaluating all possible feature subsets is an exhaustive search and thus it has high computational complexity. Until now many heuristic algorithms have been studied for solving this problem. Hyper-heuristic is a new heuristic approach which can search the solution space effectively by applying local searches appropriately. Each local search is a neighborhood searching algorithm. Since each region of the solution space can have its own characteristics, it should be chosen an appropriate local search and apply it to current solution. This task is tackled to a supervisor. The supervisor chooses a local search based on the functional history of local searches. By doing this task, it can trade of between exploitation and exploration. Since the existing heuristic cannot trade of between exploration and exploitation appropriately, the solution space has not been searched appropriately in these methods and thus they have low convergence rate.

[1] دانشجوی کارشناسی ارشد هوش مصنوعی، گروه مهندسی کامپیوتر، دانشگاه شهید باهنر، کرمان، mmontazeri@eng.uk.ac.ir
* عضو انجمن پژوهشگران جوان، دانشگاه شهید باهنر کرمان.
[2] استادیار، گروه مهندسی کامپیوتر، دانشگاه شهید باهنر، کرمان، mahdiesoleymani@yahoo.com
[3] استادیار، گروه مهندسی کامپیوتر، دانشگاه شهید باهنر، کرمان، niknafs@mail.uk.ac.ir

For the first time, in this paper use a hyper-heuristic approach to find an efficient feature subset. In the proposed method, genetic algorithm is used as a supervisor and 16 heuristic algorithms are used as local searches. Empirical study of the proposed method on several commonly used data sets from UCI data sets indicates that it outperforms recent existing methods in the literature for feature selection.

## ۱- مقدمه

استفاده از اطلاعات کلاس برای یادگیری مدل، یکی از گام‌های مهم در یادگیری بانظارت[1] و شبه نظارتی[2] است. در پایگاه داده هر نمونه متعلق به یک کلاس است که برای توصیف نمونه‌ها از تعدادی ویژگی استفاده می‌گردد. در تئوری هرچه تعداد ویژگی ها زیادتر باشد توانایی بهتری برای تفکیک کلاس‌ها و در نتیجه مدل بهتری تولید می گردد. اما در عمل تعداد نمونه های متناهی نه تنها باعث پیچیده شدن مدل می‌گردد، بلکه باعث یادگیری بیش از اندازه[3] نیز می‌شود. انتخاب ویژگی، مساله یافتن ویژگی‌های موثر از بین همه ویژگی‌ها است، به طوری که مجموعه حاصل باعث افزایش دقت و کاهش پیچیدگی گردد. این مساله در بازیابی تصاویر، سیستم‌های تشخیص حمله[4] و زیست فن آوری[5] کاربرد مهمی دارد. حل این مساله سه مزیت عمده را به همراه دارد: ۱) دقت پیش بینی را در مسایل دسته‌بندی افزایش می‌دهد. ۲) پیچیدگی را کاهش و درنتیجه سرعت پیش‌بینی را افزایش می‌دهد که این خود باعث کاهش هزینه محاسبات می‌گردد. ۳) درک بهتری از سیستم تولید داده ایجاد می‌کند. برای حل مساله انتخاب ویژگی دو مساله را بایستی تعیین کرد: معیار ارزیابی و الگوریتم جستجو. به طور کلی روش‌های انتخاب ویژگی بر اساس معیار ارزیابی به دو دسته تقسیم می‌شوند: الف) روش های مبتنی بر فیلتر[6]: این روش ها از ویژگی ذاتی داده‌ها مثل همبستگی و آنتروپی استفاده می‌کنند. در این روش‌ها به هر ویژگی یک نمره نسبت داده می شود و انتخاب ویژگی بر اساس این نمره صورت می‌گیرد [1]. این دسته از روش‌ها مستقل از الگوریتم یادگیری عمل می کنند و در نتیجه از سرعت بسیار بالایی برخوردار هستند. اما چون در تعامل با یک الگوریتم یادگیری نیستند، دقت خوبی ندارند[2]. ب)روش های مبتنی بر رپر[7]: دراین دسته از روش‌ها برای ارزیابی مجموعه ویژگی‌ها از یک الگوریتم یادگیری استفاده می‌شود و مجموعه ویژگی که بهترین دقت و کم‌ترین خطا را داشته باشد، به عنوان بهترین مجموعه ویژگی انتخاب می‌گردد. این روش‌ها نسبت به روش‌های دسته‌ی قبل از سرعت کمی برخوردار هستند، زیرا در هر مرحله از ارزیابی باید الگوریتم یادگیری آموزش یابد که این خود زمان زیادی از سیستم می‌گیرد. اما در مقابل دقت بالایی دارند[2]. تاکنون روش‌های زیادی برای حل مساله انتخاب ویژگی ارائه شده است. برخی روش-ها از نوع دسته اول هستند. در این روش‌ها از معیارهای فیلتری متفاوت مثل information gain و correlation [3,4] یا معیار mutual information [5] برای نمره دادن به ویژگی‌ها و انتخاب ویژگی‌های موثر استفاده شده است. برخی دیگر از روش‌های ارائه شده متعلق به دسته دوم هستند. به عنوان مثال در مرجع [6] از معیار دقت دسته بندی برای انتخاب ویژگی استفاده شده است. اما بیشتر روش‌ها از ترکیب این دو روش استفاده می‌کنند. برای نمونه، در مرجع [7] از الگوریتم مورچه برای انتخاب ویژگی استفاده شده است. در این روش فرمون‌هایی که هر مورچه طی سفرش بر جای می‌گذارد، بر اساس دقت دسته‌بندی ویژگی‌های کسب شده از آن مورچه است و معیار ابتکاری آن بر اساس معیارهای فیلتری است.

در روش‌های انتخاب ویژگی تعیین الگوریتم جستجوی مناسب نقش حیاتی دارد. الگوریتم‌های جستجوی زیادی بر پایه تکنیک‌های متفاوت سعی در پیدا کردن بهینه فرامحلی[8] در زمان معقول داشتند. در مراجع [8,9] از الگوریتم ژنتیک که یک الگوریتم فراابتکاری است، برای حل این مساله استفاده شده است. در این روش هر کروموزوم به صورت رشته بیت های صفر(عدم حضور ویژگی) و یک (حضور ویژگی) است. جواب بهینه مشخص کننده ویژگی های موثر است. در این روش دقت الگوریتم یادگیری به عنوان تابع هدف در نظر گرفته شده است. الگوریتم های فراابتکاری دیگر مثل جستجوی ممنوعه[9] [10]، الگوریتم پرندگان [11] و الگوریتم مورچه [12] نیز برای یافتن ویژگی های موثر استفاده شده است. برخی روش‌ها به منظور ایجاد جستجوی کارآمد و سریع از ترکیب روش‌های جستجوی متفاوت بهره برده‌اند. در مرجع [13] از ترکیب الگوریتم شبیه سازی ذوب فلزات[10]، الگوریتم وراثتی و الگوریتم تپه نوردی برای یافتن مجموعه بهینه در زمان استفاده شده است. الگوریتم‌های فراابتکاری با کاوش فضای راه حل سعی در پیدا کردن جواب بهینه دارند. ولی جستجوی محلی موثری در اطراف جواب بهینه انجام نمی‌دهند که این خود منجر به احتمال از دست دادن جواب های خوب می‌گردد.

در سال‌های اخیر استفاده از الگوریتم‌های ممتیک به عنوان الگوریتم جستجو برای حل مساله بهینه‌سازی مورد توجه محققان قرار گرفته است [14]. این روش‌ها با ارائه یک جستجوی محلی که بر اساس تغییر کوچک در راه‌حل است، توانسته اند جواب های قابل توجهی تولید کنند. در مراجع [15-18] از الگوریتم ممتیک برای یافتن ویژگی‌های موثر استفاده شده است. اما از آنجایی که جستجوی محلی وابسته به ناحیه جستجو و در نتیجه مساله است، یافتن جواب بهینه نیازمند رهیافتی است که این وابستگی را کاهش دهد که این خود منجر به تولید رهیافت جدید

ابرابتکاری گردیده است [19]. در واقع رهیافت ابرابتکاری شامل چندجستجوی محلی است که هر کدام با توجه به ناحیه جستجو اعمال می شوند. درنتیجه انعطاف پذیری مساله را به طور قابل توجهی افزایش می دهد و باعث تولید جواب‌های بهتر می گردد [20, 21]. رویکرد ابرابتکاری یک رویکرد جدید جستجو است. این رویکرد می‌تواند فضای جستجو را با به کاربستن چند جستجوی محلی، که هریک پیمایشگرهایی در همسایگی راه حل هستند، به خوبی جستجو کند. با توجه به اینکه هر ناحیه از فضای جستجو ویژگی‌های خاص خود را دارد، در مسیر جستجو بایستی یک جستجوی محلی مناسب انتخاب و در راه‌حل جاری به کاربسته شود. این انتخاب به عهده یک ناظر است. ناظر در هر زمان، انتخاب را بر اساس تاریخچه عملکرد جستجوهای محلی انجام می‌دهد. با انجام چنین تکنیکی، این رویکرد می‌تواند هم کاوش و هم بهره‌وری را به خوبی انجام دهد. الگوریتم‌های ابتکاری موجود نمی توانستند مصالحه خوبی بین کاوش و بهره‌وری داشته باشند در نتیجه فضای جستجو به خوبی جستجو نمی‌گردید و سرعت همگرایی پایینی داشتند.

در این مقاله برای اولین بار از رویکرد ابرابتکاری جهت پیداکردن ویژگی‌های موثر برای دسته‌بندی استفاده شده است. در الگوریتم پیشنهادی، هدف افزایش دقت دسته‌بندی و سرعت بخشیدن به جستجو جهت یافتن ویژگی‌های موثر است. در این روش، الگوریتم ژنتیک به عنوان یک ناظر به‌کار گرفته می‌شود و از دو نوع جستجوهای محلی استفاده می‌شود: جستجوهای محلی تپه نوردی[11] و جستجوهای جهشی[12]. در جستجوهای محلی تپه نوردی در هر مرحله سعی در یافتن راه‌حل بهتر است (بهره‌وری) ولی در جستجوهای جهشی هدف ایجاد تغییر تصادفی در راه‌حل است (کاوش). در ادامه، مقاله این گونه سازمان دهی شده است: در بخش دوم مفاهیم انتخاب ویژگی و رهیافت ابرابتکاری توضیح داده می شود. در بخش سوم الگوریتم پیشنهادی معرفی می شود. در بخش چهارم نتایج حاصل از الگوریتم پیشنهادی ارائه می گردد و در پایان در بخش پنجم به جمع بندی و نتیجه گیری می پردازیم.

## ۲- ادبیات مقاله

## ۲-۱ مساله انتخاب ویژگی

این مساله را می توان این گونه فرموله بندی کرد: اگر مجموعه ویژگی اولیه، $F$، دارای $N$ ویژگی باشد و مجموعه بهینه نهایی، x، دارای m ویژگی باشد می توان گفت انتخاب ویژگی یافتن مجموعه x به طوری که مجموعه حاصل دارای ویژگی های زیر باشد:

(1)
$m < N$

(2)
$J(x) > J(N)$

که در آن J دقت دسته‌بندی دسته‌بند مورد نظر است.

## ۲-۲ رهیافت ابرابتکاری

رهیافت ابرابتکاری اولین بار در سال ۲۰۰۰ در مرجع [23] ارائه شده است. نمودار بلوکی این رهیافت در شکل (۱) نشان داده شده است. این الگوریتم از دو لایه تشکیل شده است: لایه اول شامل تعدادی جستجوی محلی است که به آن ابتکارات سطح پایین[13]،LLH، گفته می شود. این جستجوگرهای محلی شامل قواعد یا راهبردهای متفاوتی برای حل مساله‌اند و وابسته به نوع مساله می‌باشند. اگرچه ابتکارات سطح پایین می‌توانند خود الگوریتم فرا ابتکاری باشند، ولی اغلب ابتکاراتی ساده‌ای هستند که با توجه به مساله طراحی می‌شوند. در نتیجه وابسته به مساله (مساله گرا) بوده و از یک مساله تا مساله دیگر فرق می‌کنند.

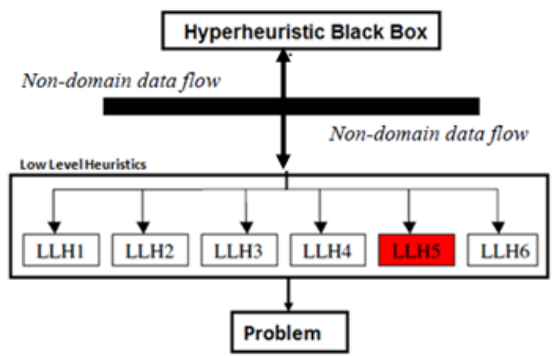

**نمودار ۱ قالب کلی رهیافت ابرابتکاری.**

دومین لایه رهیافت ابرابتکاری جعبه سیاه است که بدون هیچ مسئله خاصی طراحی می‌شود. این لایه اغلب به اطلاعاتی که به مساله وابسته نیستند (مثل اختلاف در تابع هدف ، تاریخچه اجرای هر ابتکار، حالات راه حل و....) دسترسی دارد. وظایف این لایه پذیرش یا عدم پذیرش راه حل جدید و انتخاب ابتکارات سطح پایین بعدی برای هدایت راه حل است. این لایه می‌تواند یک فرا ابتکار یا تابع انتخاب[14] باشد. این دو لایه با مرز دامنه جدا می‌شوند.

در واقع جعبه سیاه به عنوان ناظری است که انتخاب جستجوهای محلی را در هر زمان از مرحله جستجو را مدیریت می‌کند. این انتخاب با توجه به مشخصات ناحیه ای از فضای راه حل جاری در حال جستجو صورت می‌گیرد و در هر مرحله با توجه به ناحیه جستجو و تاریخچه عملکرد ابتکارات سطح پایین، ابتکار سطح پایین مناسبی انتخاب و به راه حل جاری اعمال می‌کند.

در نتیجه رهیافت ابرابتکاری با طراحی چنین سیستمی، باعث بهبودی قابل توجهی در الگوریتم‌های گذشته شده است. این رهیافت با داشتن تعدادی جستجوی محلی و اعمال مناسب آنها در هر مرحله از جستجو، باعث بهبود در الگوریتم‌های ممتیک شده است. در این مقاله هدف این است که مکانیزم جستجوی محلی موجود در الگوریتم ممتیک با رهیافت ابرابتکاری جایگزین شود و یک روش جستجوی مناسب برای مساله انتخاب ویژگی ایجاد شود.

در مرجع [24] رهیافت ابرابتکاری بر مبنای جستجوی ممنوع ارائه شده است. در الگوریتم آنها، ابتدا به هر ابتکار یک وزن نسبت داده می شود و سپس برای ارتقاء راه حل جاری، ابتکار سطح پایینی که بیش ترین وزن را داراست، انتخاب می‌شود. در صورتی که این ابتکار سطح پایین باعث بهبودی در راه حل شود وزن آن افزایش، در غیر این صورت وزن آن کاهش می‌یابد و وارد لیست ممنوعه می‌شود. در هر دو صورت جواب جاری جایگزین جواب قبلی می‌شود. در مرجع [25] الگوریتم ابرابتکاری بر مبنای SA ارائه شده است که بهبود یافته الگوریتم ارائه شده در مرجع [26] است. در این الگوریتم ابتکارات سطح پایین به صورت احتمالی انتخاب و به کار بسته می‌شوند. ابتدا به هر ابتکار سطح پایین یک وزن مناسب نسبت داده شده، سپس بر مبنای احتمال، یک ابتکار انتخاب شده و به راه حل جاری اعمال می‌شود. این ابتکار متناسب با اینکه باعث ارتقاء یا کاهش شایستگی راه حل می شود، تنبیه شده یا پاداش می گیرد. در سال 2006 یک ابرابتکار استدلال مبتنی بر حالت[15] برای مساله زمان‌بندی درسی ارائه شد[27]. این ابتکار می‌توانست بیش از اینکه مستقیماً برای یافتن جواب استفاده گردد، برای تولید راه حل‌هایی با کیفیت خوب به کار رود. نتایج حاصل از آزمایش ها نشان می‌داد که این سیستم می‌تواند به طور کارآمد و هوشمندانه در تولید جداول زمان‌بندی به کار رود. در مرجع [22] یک ابرابتکار بر مبنای الگوریتم مورچه، برای حل مساله زمان بندی پروژه[16] و مساله مسابقه سیار[17] ارائه شده است. در مرجع [28] یک رهیافت ابرابتکاری بر مبنای الگوریتم وراثتی ارائه شده است که می‌تواند به راحتی برای مسائل مختلف مورد استفاده قرار گیرد. هر فرد در جمعیت رشته‌ای از ابتکارات سطح پایین است که نشان می‌دهد که کدام ابتکار و به چه ترتیبی به کاربسته شود. این روش برای حل مساله زمان‌بندی کامیون به کار رفته که منجر به تولید جواب‌های موثر شده است. با ارتقا این روش در مرجع [29] روشی ایجاد شده است که به جای داشتن کروموزوم‌های با طول ثابت از کروموزوم‌های با طول متغیر استفاده می‌کند. در مرجع [30] یک رهیافت ابرابتکاری جدید بر مبنای رتبه‌بندی ایستای ابتکارات سطح پایین ارائه شده است. در این روش از تابع انتخابی استفاده شده که اطلاعات مربوط به عملکرد اخیر را جمع آوری می کند. تابع انتخاب شامل سه مجموعه اطلاعات است. اطلاعات اول مربوط به عملکرد هر ابتکار سطح پایین است، اطلاعات دوم مربوط به عملکرد هر جفت ابتکار سطح پایین است و اطلاعات سوم مربوط به زمان اجرای هر ابتکار سطح پایین است. در مرجع [31] از رهیافت ابرابتکاری مبتنی بر الگوریتم همسایگی متغیر[18] استفاده شده که باعث تولید جواب های خوب برای مساله مورد آزمون شده است.

## 3- الگوریتم پیشنهادی

در الگوریتم پیشنهادی، الگوریتم ژنتیک به عنوان ناظر به‌کار گرفته شده است. الگوریتم ژنتیک استفاده شده یک الگوریتم ژنتیک غیرمستقیم است که هر کروموزوم نمایش مجموعه ابتکارات سطح پایین است. نحوه کدگذاری کروزوم‌های الگوریتم ژنتیک در شکل 2 نشان داده شده است. هر ابتکار سطح پایین راه حل تغییر یافته توسط ابتکار قبلی را گرفته و بسته به نوع ابتکار به‌کاربسته شده در راه حل بهبود یا صرفاً تغییر ایجاد می‌کند. شکل 3، شبه کد الگوریتم پیشنهادی را نشان می‌دهد. هر قسمت در بخش‌های آینده توضیح داده خواهد شد.

| 10 | 9 | 1 | 6 | 2 | 3 | 8 | 11 | 13 | 5 | 7 | 10 | 12 | 14 | 9 | 4 | 15 | 16 |

باعث فراخوانی ابتکار سطح پایین 9 می گردد.

شکل 2 مثالی از نحوه کدگذاری کروموزوم الگوریتم پیشنهادی.

1. یک راه حل اولیه به صورت تصادفی ایجاد کن.($S_0$)
2. L کروموزوم اولیه با طول NLLH (تعداد ابتکارات سطح پایین) تولید کن. سپس این کروموزوم‌ها را در حوضچه ابتکارات سطح پایین قرار بده.
3. $S = S_0$.
4. به ازای هر کروموزوم، $C_i$ ($1 \leq i \leq L$) در حوضچه ابتکارات سطح پایین:
   a. ابتکارات سطح پایین را متناسب با ترتیب موجود در کروموزوم $C_i$ به S اعمال کن.
   b. راه حل تولید شده $S_i$ را ذخیره کن.
5. همه $S_i$ ها را با S مقایسه کن و در صورتی که $S_L$ بهتر از S شد قرار بده $S=S_L$.
6. عملگر انتخاب را اعمال کن و کروموزوم‌های انتخاب شده را در حوضچه ازدواج قرار بده.
7. عملگر هم‌بری را اعمال کن.
8. عملگر جهش را اعمال کن.
9. شرط پایان را چک کن. اگر شرط پایان انجام نشده برو به مرحله 4 در غیر این صورت s را به عنوان جواب بهینه ذخیره کن و خارج شو.

شکل 3 شبه کد الگوریتم پیشنهادی

## 3-1 ابتکارات سطح پایین

ابتکارات سطح پایین که جستجوی محلی نیز گفته می‌شود، باعث ایجاد تغییر در راه حل جاری می‌گردد به طوری که راه حل جدید در همسایگی راه حل جاری قرار دارد. شکل 4 نحوه کدگذاری راه حل را نشان می‌دهد. هر راه حل از رشته بیتی دودویی تشکیل شده است که یک نماینده حضور ویژگی و صفر نماینده عدم حضور ویژگی است. هر راه حل توسط معیار فیلتری پیرسون [32, 33] ارزیابی می‌گردد.

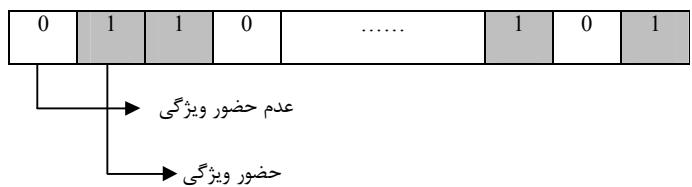

شکل 4 مثالی از کد گذاری راه حل به عنوان رشته بیتی دودویی.

ابتکارات سطح پایین استفاده شده بر دو دسته هستند: جستجوهای محلی تپه نوردی و جستجوهای جهشی. به منظور ایجاد بهره‌وری از ابتکارات سطح پایین تپه نوردی استفاده می‌گردد. این ابتکارات، در هر مرحله سعی در یافتن راه‌حل بهتر هستند و جواب تغییر یافته در صورتی پذیرفته می‌شود که بهبودی در راه حل ایجاد کند. در مسیر جستجو به منظور ایجاد کاوش و جستجوی فضاهای مختلف از جستجوهای جهشی استفاده می‌شود. این ابتکارات باعث تغییر تصادفی در راه‌حل می‌گردند. در الگوریتم پیشنهادی از 16 ابتکار سطح پایین استفاده شده است (NLLH=16). دوازده ابتکار سطح پایین استفاده شده نسخه‌های مختلف الگوریتم تپه نوردی است:

– Steepest Descent Hill Climbing (SDHC)
تمام همسایگی‌های راه حل جاری با فاصله همینگ یک را ایجاد می‌کند. از بین همه همسایگی‌ها بهترین همسایگی را انتخاب می‌کند. اگر این راه حل بهتر از راه حل جاری بود این همسایگی را جایگزین راه حل جاری می‌کند [34].

– Next Ascent Hill Climbing (NAHC)
در این روش بیت‌های راه حل جاری از با ارزش‌ترین تا کم‌ارزش‌ترین بیت بررسی می‌شوند. اگر با معکوس کردن بیت بهبودی حاصل آمد، بیت معکوس می‌شود. در حقیقت در این روش تمام راه‌حل‌های کاندید از بیت بالا به بیت پایین بررسی می‌شوند. تفاوت اصلی این روش با روش قبل در این است که در روش قبل در نهایت حداکثر یک بیت تغییر می‌کند ولی در این روش به تعداد حالت‌هایی که باعث بهبودی شده است [34].

– Davis' Bit Hill Climbing (DBHC)
این روش مانند روش قبل است با این تفاوت که ترتیب تغییر بیت‌ها مانند قبل به صورت ترتیبی از بیت با ارزش تا بیت کم ارزش نیست بلکه به صورت تصادفی جایگشتی از N محل است [35].

– Random mutation hill climbing (RMHC)

این روش به طور تصادفی یک بیت در راه حل جاری را انتخاب کرده و آن را معکوس می‌کند. اگر راه‌حل جدید بهتر یا مساوی راه‌حل جاری شد، راه‌حل جدید جایگزین راه حل جاری می‌گردد. این روش برای محل‌های تپه‌ای مفید است. زیرا در این محل‌ها مقدار تابع هدف ثابت است [34].

هر یک از چهار روش بالا به سه جستجوی محلی تعمیم داده شده‌اند که در هر یک از جستجوهای فوق یک جستجوی محلی برای کلیه بیت‌های صفر و یک و دو جستجوی دیگر یکی فقط برای بیت‌های صفر و دیگری فقط برای بیت‌های یک اعمال می‌شود. در نتیجه دوازده جستجوی محلی حاصل می‌آید. ۴ جستجوی جهشی استفاده شده عبارت است از:

- Swap dimension (SWPD)

دو بعد را به طور تصادفی انتخاب کرده و سپس این دو محل را جابه جا می‌کند.

- Dimensional Mutation (DIMM)

به طور تصادفی یک بعد را انتخاب کرده و بیت در این بعد را با احتمال ۰.۵ معکوس می‌کند.

- Hypermutation (HYPM)

هر بیت در راه حل را به طور تصادفی با احتمال ۰.۵ معکوس می‌کند.

- Mutation (MUTN)

یکبار راه حل جاری را جستجو می‌کند و هر بیت را با احتمال جهش داده شده معکوس می‌کند.

## ۳-۲ تابع شایستگی الگوریتم ژنتیک

برای ارزیابی راه حل تولید شده از دقت دسته بندی استفاده می‌شود. الگوریتم دسته بند اولین نزدیک‌ترین همسایه (1NN) به عنوان دسته بند مورد استفاده قرار گرفته و دقت حاصل به عنوان مقدار تابع شایستگی در نظر گرفته شده است.

## ۳-۳ عملگرهای الگوریتم ژنتیک

هر الگوریتم ژنتیک ۳ عملگر انتخاب، هم‌بری و جهش دارد. در روش پیشنهادی از عملگر انتخاب چرخ گردان که مبتنی بر شایستگی کروموزوم‌هاست استفاده شده است. عملگر هم‌بری استفاده شده تک نقطه‌ای است که در محل برش کروموزوم دو والد، اطلاعات دو والد جابه جا و به فرزندان منتقل می‌گردد. عمگر جهش استفاده شده یک عملگر جهش خاص هست [36]. در این عملگر تعدادی از ژن‌ها به طور تصادفی انتخاب شده و به عددی بین ۱ تا ۱۶ به غیر از عدد فعلی جهش می‌یابد.

## ۴- پیاده سازی و نتایج

الگوریتم پیشنهادی بر روی ۵پایگاه داده گرفته شده از مجموعه داده‌های UCI [37] اعمال شده است. جدول (۱) مشخصات پایگاه داده های مورد استفاده را نشان می‌دهد. جدول (۲) پارامترهای استفاده شده در الگوریتم پیشنهادی را نشان می‌دهد. به دلیل اینکه الگوریتم پیشنهادی یک جستجوی تصادفی است این الگوریتم ۱۰ بار اجرا شده و میانگین و بهترین جواب تولیدی در جدول (۳) درج شده است. تعداد ویژگی انتخاب شده توسط الگوریتم پیشنهادی در این جدول لحاظ شده است. الگوریتم دسته بند برای ارزیابی مجموعه ویژگی حاصله 1NN است. از 10 Fold Cross Validation(10 fold CV) برای ارزیابی عملکرد دسته بند استفاده شده است. چون 10 fold CV یک روال تصادفی است، بدین منظور این الگوریتم را ۱۰ بار اجرا نموده و میانگین جواب‌ها در نظر گرفته شده است (10-10 fold CV).

به منظور ارزیابی روش پیشنهادی این الگوریتم با سه روش متفاوت [38-40] مقایسه شده است. روش اول [38] یک روش مبتنی بر رپری است. روش دوم [39] یک روش فیلتری است. روش سوم [40] یک روش براساس هردو معیار رپری و فیلتری است. نتایج حاصله از مقایسه در جدول (۴) درج شده است. در هر سطر بهترین جواب پررنگ شده است. همان طور که در جدول (۴) قابل مشاهده است، الگوریتم پیشنهادی در اکثر موارد توانسته است جواب بهتری تولید کند و این  برتری روش پیشنهادی نسبت به روش های گذشته را نشان می دهد. همان طور که در جدول (۲) نشان داده شده است تعداد مراحل جستجوی الگوریتم پیشنهادی تنها ۲۰۰ بار است که نشان دهنده سرعت بالای این الگوریتم در یافتن جواب بهینه است.

**جدول (۱) مشخصات پایگاه داده ها.**

| پایگاه داده | N | تعداد کلاس | تعداد نمونه |
|---|---|---|---|
| Dermatology | 35 | 6 | 366 |
| Spectf | 45 | 2 | 349 |
| Ionoshere | 35 | 2 | 351 |
| Sonar | 61 | 2 | 208 |
| Musk | 167 | 2 | 476 |

**جدول (۲) پارامترهای الگوریتم پیشنهادی**

| احتمال هم‌بری | احتمال جهش | تعداد نسل‌ها |
|---|---|---|
| 0/7 | 0/1 | 200 |

جدول (3) بهترین و میانگین جواب های حاصل از 10 بار اجرای الگوریتم پیشنهادی (10-10 fold CV، 1NN).

| پایگاه داده | بهترین جواب، m | میانگین جواب ها، m |
|---|---|---|
| Dermatology | 0.9817, 29 | 0.9771, 27 |
| Spectf | 0.8934, 19 | 0.8823, 22.8 |
| Ionoshere | 0.9433, 12 | 0.9305, 14.5 |
| Sonar | 0.9279, 26 | 0.9034, 31.9 |
| Musk | 0.9519, 79 | 0.9407, 83.3 |

جدول (4) نتایج حاصل از الگوریتم پیشنهادی.

| پایگاه داده | روش پیشنهادی | | نتایج حاصله از منابع | |
|---|---|---|---|---|
| Dermatology | HHFS+1NN 5-10 fold CV | **0.9820, 29 (0.9774,27)** | DF-TS3-1NN+1NN 5-10-fold CV Ref.[38] | 97.28 |
| | HHFS+1NN 10-10 fold CV | **0.9817, 29 (0.9771, 27)** | DMIFS+1NN 10-10 fold CV Ref. [39] | 92.18 |
| Spectf | HHFS+1NN 5-10 fold CV | **0.8934, 22 (0.8821, 22.8)** | DF-TS3-1NN+1NN 5-10-fold CV Ref.[38] | 83.51 |
| | HHFS+1NN 10-10 fold CV | **0.8934, 19 (0.8823, 22.8)** | DMIFS+1NN 10-10 fold CV Ref. [39] | 84.79 |
| Ionoshere | HHFS+1NN 10-10 fold CV | **0.9433, 12** (0.9305, 14.5) | BCS-2+1NN 10-10-fold CV Ref.[40] | 93.3 |
| | HHFS+1NN 5-10 fold CV | 0.9419, 12 (0.9303, 14.5) | DF-TS3-1NN+1NN 5-10-fold CV Ref.[38] | **95.01** |
| Sonar | HHFS+1NN 10-10 fold CV | **0.9279, 26 (0.9034, 31.9)** | BCS-2+1NN 10-10-fold CV Ref.[40] | 89.5 |
| | HHFS+1NN 5-10 fold CV | **0.9212, 24 (0.9063, 31.9)** | DF-TS3-1NN+1NN 5-10-fold CV Ref.[38] | 90.63 |
| Musk | HHFS+1NN 5-10 fold CV | **0.9534, 79 (0.9407, 83.3)** | DF-TS3-1NN+1NN 5-10-fold CV Ref.[38] | 91.65 |
| | HHFS+1NN 10-10 fold CV | **0.9519, 79 (0.9407, 83.3)** | DMIFS+1NN 10-10 fold CV Ref. [39] | 87.34 |

## 5- نتیجه‌گیری و جمع بندی

در این مقاله برای اولین بار از رویکرد ابرابتکاری جهت پیداکردن ویژگی‌های موثر برای دسته‌بندی استفاده شد. در الگوریتم پیشنهادی، هدف افزایش دقت دسته‌بندی و سرعت بخشیدن به جستجو جهت یافتن ویژگی‌های موثر بود. در این روش، الگوریتم ژنتیک به عنوان یک ناظر به‌کار گرفته شد که انتخاب مناسب ابتکارات سطح پایین را کنترل کند. هر ناحیه از فضای جستجو ویژگی‌های خاص خود را دارد و هر ناحیه با ناحیه دیگر متفاوت است. با به کاربستن ابتکارات متفاوت که جستجوگرهای محلی هستند، الگوریتم پیشنهادی توانست فضای جستجوی را به خوبی جستجو کند. در این روش از دو نوع ابتکارات سطح پایین جستجوهای محلی تپه نوردی و جستجوهای جهشی استفاده شد. هدف از جستجوهای محلی تپه نوردی افزایش بهره‌وری است و جستجوهای جهشی هدف کاوش فضای جستجو است. ایجاد مصالحه بین بهره‌وری و کاوش ویژگی مهم

هر جستجوگر ابتکاری است که در الگوریتم پیشنهادی این مصالحه به خوبی انجام شده است. روش پیشنهادی که یک روش رپر-فیلتری است که بر روی پایگاه داده های متفاوت گرفته شده از UCI به کار بسته شد و با انواع روش‌های مختلف انتخاب ویژگی، رپری، فیلتری و رپر-فیلتری مقایسه شد. نتایج مقایسه نشان دهنده کارامد بودن روش پیشنهادی در مقایسه با سایر روش‌های موجود برای انتخاب ویژگی است.

## تقدیر و تشکر

## منابع

---

[1] Supervised learning

[2] Semi-supervised learning

[3] Over fitting

[4] Intrusion detection system

[5] Bioinformatics

[6] Filter based method

[7] Wrapper based method

[8] Global Optimum

[9] Tabu Search

[10] Simulated Annealing

[11] Hill climbers local searches

[12] Mutational local searches

[13] Low Level Heuristic (LLH)

[14] Choice Function

[15] Case Base Reasoning

[16] Project Presentation Scheduling Problem

[17] Travelling Tournament Problem

[18] Variable Neighborhood Search